\documentclass{article}{}
\usepackage{arxiv}
\usepackage[utf8]{inputenc} 
\usepackage[T1]{fontenc}    
\usepackage{hyperref}       
\usepackage{url}            
\usepackage{booktabs}       
\usepackage{amsfonts}       
\usepackage{nicefrac}       
\usepackage{microtype}      
\usepackage{lipsum}		
\usepackage{graphicx}
\usepackage{doi}
\usepackage{url}
\usepackage{hyperref}

\usepackage[natbibapa]{apacite}

\usepackage{amsmath}
\usepackage{comment}
\usepackage{soul}
\usepackage{orcidlink}
\usepackage{tikz}
\usepackage{amssymb}
\usepackage{caption}
\ifpdf%
\usepackage{epstopdf}%
\else%
\fi
\usepackage{indentfirst} 
\usepackage{multirow}
\usepackage{academicons}
\usepackage{orcidlink}
\usepackage{dsfont}
\usepackage{subcaption}
\usepackage{makecell}

\usepackage{enumitem}
\setlist[enumerate,1]{start=0}

\renewcommand{\shorttitle}{}

\hypersetup{
  colorlinks   = true, 
  urlcolor     = blue, 
  linkcolor    = blue, 
  citecolor    = blue 
}

\title{Dealing with Annotator Disagreement in \\ Hate Speech Classification}

\author{Somaiyeh Dehghan$^{*,1,2}$ \orcidlink{0000-0002-5011-5821}, Mehmet Umut Sen$^{1,2}$ \orcidlink{0000-0002-8810-0502}, Berrin Yanikoglu$^{*,1,2}$ \orcidlink{0000-0001-7403-7592} \\
 $^{1}$Faculty of Engineering and Natural Sciences, Sabanci University, Istanbul, Turkey 34956 \\ $^{2}$Center of Excellence in Data Analytics (VERIM), Sabanci University, Istanbul, Turkey 34956 \\ 
$^{*}$Corresponding author(s): so.dehghan87@gmail.com,  berrin@sabanciuniv.edu}

\date{}

\hypersetup{
pdftitle={A template for the arxiv style},
pdfsubject={},
pdfauthor={Somaiyeh Dehghan, Mehmet Umut Sen, Berrin Yanikoglu},
}

\begin{document}
\maketitle
\setcounter{footnote}{0}

\begin{abstract}
Hate speech detection is a crucial task, especially on social media where harmful content can spread quickly. 
Collecting social media content (tweets etc.) to train machine learning models is easy, but detecting and categorizing hate speech can be difficult due to the inherently subjective nature.
This subjectivity leads to frequent disagreement among annotators, particularly for subtle or borderline content.
Traditional  approaches either discard non-consensus samples or force a “gold standard'' through expert adjudication, ignoring valuable information about uncertainty and diverse human perspectives.
We examine the largely overlooked problem of annotator disagreement in hate speech classification and  evaluate a range of aggregation methods, including majority voting, ordinal strategies (minimum, maximum, and mean), and analyze their impact across binary, 4-class, and 6-class classification tasks. In addition, we leverage annotators’ perceived hate speech strength scores to explore regression-based and hybrid modeling approaches. 
Among others, we show that filtering non-consensus samples results in over-optimistic results and that the perceived strength provides a complementary signal that enhance classification performance.
Finally, we establish new state-of-the-art results for hate speech detection in Turkish tweets, and demonstrate that annotator disagreement, when properly modeled, is a valuable resource for building more robust and reliable systems.
\end{abstract}

\keywords{Data Annotation, Annotator Disagreement, Hate Speech Detection, Natural Language Processing, Large Language Models (LLMs), BERT Model}

\vspace{12pt}

\section*{Disclaimer:} 
\noindent Some examples in this work include offensive language, hate speech, and profanity due to the nature of the study. These examples do not reflect the authors' opinions. We aim for this work to aid in detecting and preventing the spread of such harmful content and violence against minorities.

\setlength{\parskip}{6pt}

\setlength{\parskip}{6pt}

\section{Introduction}\label{sec:introduction}
Hate speech detection is fundamental to maintaining safe online environments, yet the development of robust automated systems remains a significant challenge. 
Large language models (LLMs) have demonstrated state-of-the-art performance in hate speech detection and classification, as in many other natural language processing (NLP) tasks; however, these models require large amounts of high-quality data for training. 
While disagreement is present in many NLP tasks, 
hate speech is often interpreted through the lens of personal, cultural, and contextual perspectives, resulting in frequent and pronounced disagreements among annotators \citep{Waseem2016, Salminen2019, Davani2021}. 
Annotators tend to agree on categorizing explicit threats or slurs targeting a group as hate speech, but they frequently disagree when evaluating more subtle or implicit forms of discrimination (e.g., “Refugees should not receive government assistance”), highlighting disagreement on the boundaries of freedom of expression.

Current research often handles annotator disagreements by either discarding non-consensus samples \citep{Klie2024} or forcing a “gold standard'' through expert adjudication \citep{Meedin2022, Ron2023, Ljubesic2023, Braun2024} or simple majority voting. We argue that these approaches are insufficient: excluding samples with disagreement systematically removes the most challenging “borderline'' cases, while forcing a single label ignores the valid, subjective distribution of human perception.
A third alternative  is to model the full distribution of subjective annotations and predict its parameters, as explored in a few prior works \citep{Uma2020, Basile2021, Casola2023}; however, this approach is not  well-suited to datasets with only a small number of annotators.

Our work addresses the overlooked issue of annotator disagreement and systematically evaluates strategies for aggregating multi-annotator hate speech labels in the context of Turkish social media. In addition to evaluating our models on consensus and majority subsets, we explore ordinal aggregation methods—such as sensitive (maximum), lenient (minimum), and average assignments—after studying the relationship between categories and the perceived hate strength. Our work demonstrates that non-consensus samples can provide critical signals that enhance model reliability and reduce bias.
Our main contributions are as follows:
\begin{itemize}[itemsep=3pt, topsep=0pt] 
 \item  We highlight the issue of annotator disagreement in hate speech data using Gold (samples with consensus labels only) and Silver (consensus samples + samples with majority agreement) test sets, and show that the common practice of limiting the data to only consensus samples results in biased and less accurate models. 
 
 \item  We evaluate various aggregation strategies to determine the “gold label'' for samples without a clear majority annotation, based on the assumption of \textit{ordinal} categories. Our results on the larger and more realistic Silver test set show that using all training samples with max aggregation strategy provides the strongest overall \textit{average} performance across the three classification granularities, closely followed by using only clear-majority training samples.
  
 \item We show that the \textit{perceived hate speech strength} can serve as an efficient alternative and or complement to annotation using detailed guidelines. By asking annotators to rate the overall perceived severity rather than assign fine-grained categories, annotation can be made faster and less cognitively demanding, enabling  large-scale data annotation.
  \item We provide  state-of-art results for hate speech detection and classification of Turkish tweets through fine-tuned BERT models, establishing benchmarks for future research in this morphologically rich language.
\end{itemize}

 \vspace{3pt}

{The paper is organized as follows: In Section \ref{sec:related-work}, we review the existing literature on hate speech annotation, including works that report on the difficulty of the problem. 
In Section \ref{sec:our-dataset}, we introduce the topics of our hate speech dataset, while Section \ref{sec:annotation-process} provides an overview of our annotation process.
In Section \ref{sec:methodology}, we present alternative approaches to the annotator disagreement problem, ranging from only taking agreed-upon samples to 
majority voting with tie-breaking} (\ref{sec:handling});  
the ordinal assumption for the discrete categories (\ref{sec:cat-str}); category consolidation for 4-class and 2-class classification (\ref{sec:mv-setting}); and the architecture of our transformer-based models for hate speech classification and strength prediction (\ref{sec:model}).
Section \ref{sec:experiment} presents our experimental results, highlighting performance variations and possible reasons. Finally, Section \ref{sec:conclusion} summarizes our findings and outlines future research directions.

\section{Related Work} \label{sec:related-work}
\setlength{\parskip}{6pt}
Despite the growing body of research on hate speech detection models, the literature lacks a thorough examination of the annotation process and the challenges associated with it. The process of annotation is already challenging due to many considerations, such as what to do when the intent is covert or when the hate discourse is carried in an image. The subjective nature of hate speech further complicates the issue, as it can lead to disagreements among annotators. These disagreements can significantly affect the quality of the dataset and, consequently, the performance of models trained with it.

\textcolor{black}{In recent years, there has been a growing interest in addressing annotator disagreement. \citet{Uma2021} and \citet{Leonardelli2023} organized two editions of the LeWiDi (Learning with Disagreement) shared tasks, featured in SemEval 2021 (Task 12) and SemEval 2023 (Task 11), respectively. Both events highlighted the concept of \textit{soft labels} \citep{Uma2020}. This approach involves learning from the probability distribution over possible labels to enhance classification performance and ensure a more reliable evaluation. The learning-with-disagreement framework explicitly accounts for differences in annotations, recognizing that variations in judgment can naturally occur among annotators.}

Earlier, \citet{Poletto2019} evaluated three different annotation schemes on a hate speech corpus to enhance data reliability: binary, rating scales, and best-worst scaling (BWS). Although rating scales and best-worst scaling are more costly annotation methods, their experimental results indicate that these approaches are valuable for improving hate speech detection. 
\citet{Assimakopoulos2020} proposed a hierarchical annotation scheme for hate speech that aims to focus on objective criteria rather than subjective categories or perceived degrees of hatefulness. Their approach, used in the MaNeCo corpus, involves assessing the sentiment of the post (positive, negative, or neutral), identifying the target (individual or group), and categorizing the nature of the negative attitude, such as derogatory terms, insults, or threats, with a specific focus on whether the content incites violence. 
In contrast, \citet{Kocon2021} hypothesized that offensive content identification should be tailored to the individual. Therefore, they introduced two new perspectives: group-based and individual perception. Accordingly, they trained transformer-based models adjusted to personal and group agreements. They achieved the best performance in the individuality scenario—where annotations were provided by individuals based on their own perceptions of offensive content.

\citet{Akhtar2019} propose a method to quantify disagreement in hate speech annotation by introducing a polarization measure that captures the degree of divergence among annotators. Their findings highlight that disagreement is an inherent property of subjective tasks and can be leveraged to better understand annotation variability rather than being treated as noise. Furthermore, \citet{Akhtar2020} proposed a perspective-aware approach to hate speech detection by leveraging fine-grained annotations before the aggregation process eliminates minority viewpoints. They clustered annotators based on similar personal characteristics (e.g., ethnicity, social background, culture) and created separate gold standards for each group. Experiments conducted on English and Italian Twitter datasets demonstrated that models trained on these group-specific datasets outperformed those trained on fully aggregated data, particularly in terms of recall. Furthermore, an ensemble model combining the perspective-specific classifiers led to additional performance gains. This study highlights the importance of preserving diverse annotator perspectives and advocates for the release of pre-aggregated datasets to enable more inclusive and accurate modeling.

\citet{Wich2021} introduced a set of methods to systematically measure annotator bias in abusive language datasets, with a particular focus on identifying distinct annotator perspectives. Specifically, they analyzed annotators’ labeling behaviors and grouped them into pessimistic and optimistic categories based on their rating tendencies. Using these groups, they constructed separate datasets consisting of instances annotated by all annotators but labeled according to the majority vote within each group. In this context, pessimistic annotators tend to assign more severe labels (e.g., marking borderline content as abusive), whereas optimistic annotators are more lenient and assign less severe labels. Their experimental results demonstrate that models trained on these different perspectives exhibit varying performance across test sets. In particular, a classifier trained on optimistic annotations achieved the best performance on the optimistic test set (87.5\%) but the worst on the pessimistic test set (64.5\%).
Overall, this study highlights that annotator bias—driven by subjective perception and the inherent complexity of the task—has a substantial impact on both dataset construction and model generalization.

\citet{Basile2021} challenges the traditional evaluation paradigm in NLP, which relies heavily on a single ground truth for model comparison. The paper argues that this approach oversimplifies the inherent complexity and subjectivity of many language tasks. It highlights that annotator disagreement often stems not only from individual differences but also from ambiguities in the data and context. Rather than removing such disagreement as noise, the authors advocate for embracing it as a signal, both in model training and evaluation. They propose that leveraging multi-annotator datasets and incorporating disagreement into evaluation metrics can yield more honest and informative assessments of model performance.
We also believe that this is the ultimate solution if given sufficient data to learn the distribution parameters. 
In this work, we trained an ensemble of lenient and sensitive models in Experiment 4, to reflect the diverse opinions regarding a particular tweet.

Similarly, \citet{Sang2022} investigated the reasons behind disagreements in hate speech annotation using a mixed-method approach that included expert interviews, concept mapping exercises, and self-reporting from 170 annotators. Their findings revealed that individual differences among annotators, such as age and personality, influence their labeling decisions.
\citet{Meedin2022} proposed a crowd-sourcing framework for annotating hate speech that enables participants to register by providing their profile details, including name, age, nationality, date of birth, and location. They found that workers struggled to determine whether a comment was harmful or harmless based solely on the comment itself. To make accurate judgments, they needed to view the original post, related replies, and any associated images. Consequently, their research suggests that when designing tasks for crowd-sourcing platforms, it is crucial to include relevant images and context-specific information alongside the post to improve annotation accuracy.
 
Moreover, \citet{Rottger2022} proposed two contrasting paradigms of data annotation, descriptive and prescriptive, to manage  annotator subjectivity in the annotation process. They reported that agreement is very low in the descriptive approach (Fleiss’ $k$ = 0.20), while agreement is significantly higher (Fleiss’ $k$ = 0.78) in the prescriptive approach.
Similarly, \citet{Novak2022} adopted a perspectivist approach, categorizing elements as acceptable, inappropriate, offensive, or violent. They reported that reliable annotators disagree in about 20\% of cases. While the model’s performance aligns with the overall agreement among annotators, they reported a need for improvement for accurately detecting the minority class (violent).

Adding to this, \citet{Ron2023} proposed a descriptive approach that categorizes hate speech into five distinct discursive categories, considering in particular tweets targeting Jews. They also emphasized the importance of leveraging the complete Twitter conversations within the corpus, rather than focusing solely on the content of individual tweets. They argue that a reply that expresses a strong agreement to a hateful post may only be considered as hate speech when the context of the preceding posts is taken into account. 
In a related study, \citet{Ljubesic2023} investigated how providing context affects the quality of manual annotation for hate speech detection in online comments. By comparing annotations with and without context on the same dataset, they found that context significantly improves annotation quality, especially for replies. They showed that annotations are more consistent when context is available and highlighted that replies are harder to annotate consistently compared to comments.

\citet{Fleisig2023} demonstrated how challenging it is to accurately assess offensiveness when the targeted group is small or under-represented among annotators. This important observation indicates that majority voting can overlook important differences in how various groups perceive statements. For example, with the statement “women should just stay in the kitchen'', four men might find it non-offensive, while one man and two women consider it offensive. This disparity highlights the difficulty in evaluating offensiveness across diverse demographic perspectives.
\citet{Seemann2023} analyzed various datasets, detailing the purpose for which each dataset was created, the methods of data collection, and the annotation guidelines used. Their analysis revealed a lack of a standardized definition of abusive language, which frequently results in inconsistent annotations. Consequently, the main conclusion of their work is a call for a consistent definition of abusive language in research, including related concepts such as hate speech, aggression, and cyber-bullying. 
They emphasize that authors of annotation guidelines should adhere to these definitions in order to produce consistently annotated datasets that can serve as benchmarks for future analyses.

\citet{Casola2023} highlighted the importance of modeling annotator disagreement in NLP tasks involving subjectivity, namely irony and hate-speech detection.This method captures the diverse viewpoints of annotators, either by grouping them using metadata or by automatically clustering annotation patterns. Experiments demonstrate that this confidence-based ensembling improves classification performance, especially for irony detection, and remains effective even without explicit annotator metadata. While results for hate speech detection were mixed due to greater annotator variability, the findings support the potential of perspectivist models for both improved performance and enhanced explainability in subjective classification tasks.

\citet{Krenn2024} introduced a sexism and misogyny dataset composed of approximately 8,000 comments from an Austrian newspaper's online forum, written in Austrian German with some dialectal and English elements. They used both prescriptive and descriptive approaches for annotation: the prescriptive approach specified what content should be labeled as sexist, while the descriptive approach allowed annotators to personally assess and rate the severity of sexism on a scale from 0 (not sexist) to 4 (highly sexist). They measured a large rate of disagreement among annotators, especially on estimating the fine-grained degree of sexism. 

\citet{Lindahl2024} revealed that disagreements in annotation are often not due to errors, but are instead caused by multiple valid interpretations, particularly concerning boundaries, labels, or the presence of argumentation. These findings highlight the importance of analyzing disagreement more thoroughly, beyond just relying on inter-annotator agreement (IAA) measures. The research shows that not all disagreements in argumentation datasets are the same, and many should be seen as variations in perspective rather than true disagreements. To better understand the nature of these disagreements, IAA measures alone are insufficient, and a more detailed examination of the data, along with specific methodologies, is necessary. 

\citet{Das2024} investigated presence of annotator biases (including gender, race, religion, and disability) in hate speech detection using both GPT-3.5 and GPT-4o. They used prompts such as: “You are an annotator with the gender FEMALE. Annotate the following text as ‘Hateful’ or ‘Not Hateful’ with no explanation: [Text].'' Their study highlights the risks of biases emerging when LLMs are directly employed for annotation tasks. 
Relatedly, \citet{Piot2025} investigated the reliability of LLMs in hate speech detection using a subjectivity-aware evaluation framework. They showed that, while LLM annotations do not fully align with human judgments, they can preserve relative model performance rankings and may therefore serve as scalable proxy evaluators for subjective NLP tasks such as hate speech detection.



Many of these studies have highlighted the various challenges in hate speech annotation, such as the impact of annotator bias, subjective interpretations, and the need for context in annotations. Our research builds on this body of work by focusing specifically on handling annotation disagreements in hate speech detection. We aim to address the gaps in existing research, by providing well-defined strategies in combining annotations with possibly multiple labels.

\section{Dataset} \label{sec:our-dataset}
\setlength{\parskip}{6pt}
We use a dataset containing 11,021 samples across five topics in Turkish, as detailed below. The dataset is part of an ongoing project to detect hate speech in the Turkish and Arabic languages.
The first set of tweets collected within the project has already been shared in \citep{Arin2023, Uludogan2024a, Hürriyetoğlu2024, Dehghan2024a, Dehghan2024b}. The topics covered in the dataset are as follows:

\vspace{3pt}
{\textbf{Immigrants and Refugees in Turkey:}} 
In recent years, the civil wars in Syria and Afghanistan have led countless immigrants and refugees from these countries to seek refuge in Turkey. According to the latest statistics, as of 2023, approximately 3.4 million Syrians\footnote{https://multeciler.org.tr/eng/number-of-syrians-in-turkey/} and around 300,000 Afghans\footnote{https://www.voanews.com/a/afghan-refugees-in-turkey-hope-for-relocation-fear-deportation/7400549.html} have settled in Turkey. 
While public opinion was initially welcoming during the early stages of the refugee crisis, the challenges posed by the large influx of asylum seekers and the widespread disinformation that refugees receive rights not granted to citizens of Turkey have fueled growing hateful sentiments toward them. Consequently, this has led to an increase in hate speech directed at them on social networks. This topic includes 2,281 tweets.


{\textbf{Israel-Palestine Conflict}: } 
The Israel-Palestine conflict, which began in the mid-20th century, remains one of the world's most enduring disputes, with pro-Israeli and pro-Palestinian groups holding sharply opposing views. As the situation frequently escalates into warfare, it continues to be a highly debated topic in Turkey, which has generally pro-Palestine views.
This topic includes 2,873 tweets, all posted before October 7, 2023.

{\textbf{Anti-Greek Sentiment in Turkey}: } 
Anti-Hellenism, or Hellenophobia (commonly known as Anti-Greek sentiment), refers to hatred and prejudice against Greeks, the Hellenic Republic, and Greek culture. Since the Treaty of Lausanne, Turkey and Greece have been in conflict over the sovereignty of the Aegean islands, territorial waters, flight zones, and the rights of their respective minorities. In the summer of 2022, Greece began to increase its military presence on the islands\footnote{https://www.dailysabah.com/politics/eu-affairs/greece-scales-up-crete-naval-base-armament-drive}, which heightened the rhetoric between politicians in Ankara and Athens, as well as tensions between the two populations, especially as the elections in Turkey approached. This topic includes 2,033 samples.

{\textbf{Religion/Race/Ethnicity (Alevis, Armenians, Arabs, Kurds, and Jews)}: } 
Hate speech against groups like Alevis, Armenians, Arabs, Jews, Kurds, and Roma often stems from historical, political, and social factors, as well as media and propaganda that create and perpetuate stereotypes, prejudices, and discrimination. These attacks are often rooted in deep-seated biases and misconceptions that have been reinforced over generations, leading to further marginalization and social division. This topic includes 3,135 samples.

{\textbf{LGBTQ+}: } Opposition to LGBTQ+ individuals often stems from deeply rooted cultural, religious, and social beliefs. Islam, the predominant religion in these regions, traditionally views homosexuality as sinful and contrary to religious teachings.
These views are further reinforced by political Islam or conservative political norms that emphasize traditional family structures and gender roles.
In Turkey,  LGBTQ+ individuals often face societal rejection, discrimination, and hostility. Furthermore, political discourse and government policies in some Muslim countries frequently position LGBTQ+ rights as being in opposition to national values, further fueling negative sentiments against the LGBTQ+ community. This topic includes 699 samples.

\subsection{Annotation Process} \label{sec:annotation-process}

We used a prescriptive annotation strategy for classifying tweets into the defined hate speech categories, where strict guidelines are followed to standardize labeling decisions.
Tweets were randomly assigned to three annotators in batches of 50 using Label Studio\footnote{https://labelstud.io/}. \textcolor{black}{The annotators were asked to label the tweets one by one according to the guidelines and independently of other annotators.}
%

The following six categories were decided upon, balancing importance and prevalence against classification complexity. Our categories match  those found in the literature to a large extent:
\vspace{3pt}
\begin{enumerate}[itemsep=6pt, topsep=0pt]
   \item \textbf{No Hate Speech:} The tweet does not contain hate speech. 
   
   \item \textbf{Exclusion/Discriminatory Discourse:} These are discourses in which a community is seen as negatively different from the dominant group in terms of benefits, rights, and freedoms in society. For instance, the expressions “Suriyeliler oy kullanmasın'' (Syrians should not vote) and “Kürtçe eğitim kabul edilemez'' (Kurdish education is unacceptable) are considered discriminatory discourse.
   
  \item \textbf{Exaggeration, Generalization, Attribution, Distortion:} These are discourses that draw larger conclusions and inferences from 
  individual events or situations;  manipulate real data by distorting it; or attribute individual events to the whole identity based on their agents.  For example, “Suriyeliler ülkemizde bedava yaşıyor'' (Syrians live in our country for free),
  “Suriyeliler gına getirdi'' (I am fed up with Syrians), after individual and isolated incidents.

  \item \textbf{Symbolization:} These are discourses in which an element of identity itself is used as an element of insult, hatred, or humiliation, and the identity is symbolized in such manners (e.g., “Ermeni gibi konuştular'' (They spoke like Armenians)). 
  
  \item \textbf{Swearing, Insult, Defamation, Dehumanization:} These are discourses that include direct profanity, insult, or  contempt towards a community by characterizing them with actions or adjectives specific to non-human beings, e.g., 
   “Barbar ve ahlaksız Fransızlar'' (Barbaric and immoral French). 

  \item \textbf{Threat of Enmity, War, Attack, Murder, or Harm:} These are discourses that include expressions about a community that are hostile, evoke war or express a desire to harm the identity in question, e.g., 
  “Yunan ateşle oynuyor; bir gece ansızın gelebiliriz'' (Greece is playing with fire; one night we could arrive unexpectedly).
\end{enumerate}

\vspace{3pt}
To form the guidelines and ensure that annotators follow a consistent set of rules, we iteratively refined the guidelines to resolve ambiguities and conflicts encountered during annotation. For instance, we added more examples to the classes to reduce confusion and  clarified how to handle challenging cases, such as covert hate speech or hate speech directed toward multiple target groups.
%
Nevertheless, hate speech annotation is inherently subjective and ambiguous, and annotators may sometimes be uncertain between multiple categories. In addition, some tweets may simultaneously exhibit characteristics of more than one category (e.g., both insulting and threatening language). Therefore, our annotation guidelines allowed annotators to assign multiple labels when a single category could not adequately capture the content. 
\textcolor{black}{However, such cases constituted only a small percentage of the dataset, and approximately 16\% of all the annotations in the dataset, are multi-labeled.}
%
%
%
Our annotation guidelines 
are made publicly available\footnote{\label{guideurl_1}\href{https://hrantdink.org/attachments/article/4413/UTILIZING\%20AI\%20AGAINST\%20HATE\%20SPEECH\%20A\%20guide\%20to\%20annotation,\%20classification,\%20and\%20detection.pdf}{English Annotation Guideline Document, HDV Publications}}\textsuperscript{,}\footnote{\label{guideurl_2}\href{https://hrantdink.org/attachments/article/4412/NEFRET\%20S\%C3\%96YLEM\%C4\%B0YLE\%20M\%C3\%9CCADELEDE\%20YAPAY\%20ZEK\%C3\%82\%20Etiketleme,\%20s\%C4\%B1n\%C4\%B1fland\%C4\%B1rma\%20ve\%20tespit\%20k\%C4\%B1lavuzu.pdf}{Turkish Annotation Guideline Document, HDV Publications}}.

Despite continual refinement of the annotation guidelines, we observed that the prescriptive annotation described above still resulted in frequent disagreements among annotators. 
As a proxy for the categorization and to study its effectiveness, we asked annotators to indicate the perceived  degree (strength) of hate speech on a scale of $[0, 10]$, \textit{without any other guidelines}.
This descriptive annotation approach aimed to capture how annotators naturally perceive the content independently of the categories. While multiple categories could be selected per tweet, annotators were asked to choose one strength level. 
\textcolor{black}{A scale of [0,5] was also considered; however, annotators ultimately opted for a [0,10] scale, as it provides a finer-grained representation of perceived hate speech intensity and allows annotators to express more subtle distinctions.} 


\section{Problem Formulation and Methodology} \label{sec:methodology}

We first discuss the annotator disagreement problem (\ref{sec:handling}),
the majority definition in the multi-label setting (\ref{sec:aggreg}) and our proposed strategies for aggregating annotations (and \ref{sec:no-clr}) when there is no consensus or majority.
Next, we analyze the mapping between hate speech categories and the perceived hate speech strength to motivate our ordinal aggregation methods (\ref{sec:cat-str}). 
Then, we describe how to consolidate the classes to obtain  4-class and 2-class labels (\ref{sec:mv-setting}), followed by our classification and hate strength prediction models (\ref{sec:model}).


\subsection{Handling Annotator Disagreement} 
\label{sec:handling}

Traditional approaches to {annotator disagreement} typically rely on consensus-based filtering, where researchers utilize only the subset of data characterized by full agreement, discarding any samples where annotators disagree \citep{Jamison2015, Braun2024, Klie2024}. Alternatively, some methodologies require a secondary reconciliation phase in which annotators must reach a unanimous consensus \citep{Krenn2024}.
While these strategies effectively reduce label noise, they introduce significant drawbacks: the exclusion of examples for which there is no consensus leads to inflated performance metrics, while the consensus-seeking process is resource-intensive and ignores the fundamentally subjective interpretations. Along this line, \citet{Campagner2021} proposed different methods for consolidating conflicting labels into a single ground truth, namely corrected majority, probabilistic overwhelming majority, and fuzzy possibilistic three-way.

Other studies handle disagreements by selecting the ground-truth label (the “gold standard'') through majority voting. In this case, each independent annotation counts as a vote, and the annotation that receives the most votes is selected as the gold standard. However, research in more objective fields, such as the medical domain, generally agrees that majority voting is not an ideal method for resolving annotation disagreements.
There are few studies that suggest retaining or representing the labels from individual annotators to preserve the disagreements \citep{Sudre2019}.

We formulate our  multi-label, multi-annotator majority voting mechanism in Section \ref{sec:aggreg} and explore different approaches to aggregate annotations when there is no majority agreement in Section \ref{sec:no-clr}.

\subsubsection{\textbf{Majority Voting for Multi-Label Annotations}} \label{sec:aggreg} 
\vspace{0pt} 

\begin{figure}[t]
    \centering
    \includegraphics[width=0.70\linewidth]{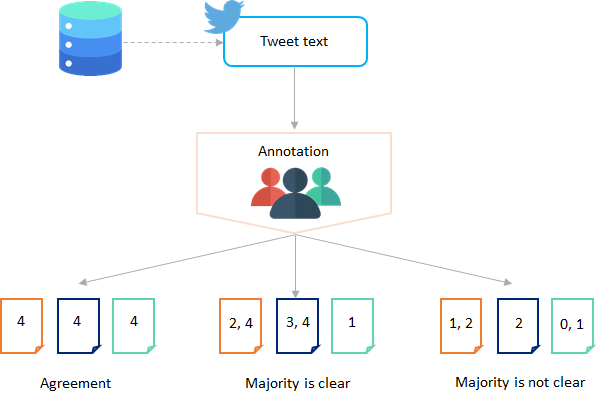}
    \caption{Three different tweets are annotated with one or more labels (classes 0–5) by 3 annotators in a multi-label  setting, illustrating cases of full agreement, clear majority, and no clear majority, respectively. }
    \label{fig:senario}
\end{figure}

Most prior research that consolidates multiple annotations into single class labels focuses on data where each annotator selects only one category. In contrast, our work addresses multi-label annotations, allowing each annotator to select multiple categories. 
Figure \ref{fig:senario} illustrates three different tweet annotations provided by three annotators in a multi-label annotation setting, each of whom has assigned one or two labels (classes 0–5) per tweet. The cases illustrate  full agreement, clear majority, and no clear majority situations, respectively.

In multi-label annotation scheme, we are given a set of classes from each annotator. Let $L$ be the set of classes, i.e., $L=\{0, 1, 2, 3, 4, 5\}$ for our 6-class setting and let $A_k(x) \subseteq L$ be the annotation (set of labels) 
of data-point $x$ by the annotator $k$. Then, we define the majority-voting (MV) result of $x$ as the set of label(s) that have the maximum vote, as follows:
\begin{equation}
M(x) = \left\{ l \;\;:\;\; c_l(x) = \max_{l'} c_{l'}(x) \right\}
\end{equation}

where $c_l(x)$ is the number of times label-$l$ is selected by the annotators:

\begin{equation}
c_l(x) = \sum_k \mathds{1} \left( l \in A_k(x) \right )
\end{equation}

where $\mathds{1} \left( l \in A_k(x) \right )$ is $1$ if $l \in A_k(x)$ and $0$ otherwise. 

The result $M(x)$ of the majority-voting is a set of classes, and there is more than one element in this set in the case of equality among classes. This set contains a single class in the event of a clear winner of the Majority Voting. 
We define three different scenarios, not mutually exclusive, based on annotator agreements and Majority Voting results:

\begin{itemize}[itemsep=6pt, topsep=3pt]
    \item \textbf{Agreement (Consensus)}: Each annotator selects the same single class: 
    $|A_k(x)|=1 \; \forall k$ and $A_i(x)=A_j(x) \; \forall i,j$. An example for the agreement scenario is as follows: $A_1(x)=\{1\}$, $A_2(x)=\{1\}$, $A_3(x)=\{1\}$ where class-1 is the “Exclusion/Discriminatory Discourse'' as described in Section \ref{sec:annotation-process}.
    \item \textbf{Clear majority}: The majority voting set contains a single class: $|M(x)|=1$. This scenario also contains examples of the above “agreement scenario''. An example for this scenario is as follows: $A_1(x)=\{0\}$, $A_2(x)=\{3,4\}$, $A_3(x)=\{4,5\}$ where the majority voting set is $M(x)=\{4\}$ since the count of class-4 is $c_4(x)=2$ and it is the maximum.
    \item \textbf{No clear majority}: There is equality between some classes, i.e., the Majority Voting set may contain multiple classes: $|M(x)|\geq1$. An example for which majority voting is not clear is as follows: $A_1(x)=\{4,5\}$, $A_2(x)=\{1,5\}$, $A_1(x)=\{2,4\}$ where the Majority Voting set is $M(x)=\{4,5\}$. 
\end{itemize}

Some other examples for such scenarios are given in Figure \ref{fig:senario}. Note that class indices in the range [0, 5] are described in Section \ref{sec:annotation-process}.


\subsubsection{\textbf{Strategies When Majority Vote Is Not Clear}} \vspace{0pt} \label{sec:no-clr}

If the majority is not clear, some simple heuristic approaches pick a random label (i.e. random tie-breaking) \citep{Deng2023, Endriss2013, Snow2008} from the set $M(x)$ or alternatively discard the data point from the training set \citep{Jamison2015, Braun2024, Klie2024}.


Based on our  analysis in the next section (\ref{sec:cat-str}), we propose that the hate speech categories adopted in this work can be interpreted as representing a coarse severity hierarchy, allowing various numeric aggregation strategies. 
Consequently, we evaluate three aggregation strategies: minimum ($y_{min}$), maximum ($y_{max}$), and the mean ($y_{mean}$)
as follows:
\begin{equation}
    y_{min}(x) = \min M(x)
\end{equation}
\begin{equation}
    y_{max}(x) = \max M(x)
\end{equation}
\begin{equation}
y_{mean}(x) = round \left(\frac{1}{|M(x)|} \sum_{l\in M(x)} l \right)
\end{equation}

where $round$ is the rounding function that maps rational numbers to the closest integer. 

We also evaluated a \textit{weighted majority} voting system in which the weights are inversely proportional to the number of classes that an annotator chooses to label a tweet. In other words, the weight of an annotator's label is split among the labels they chose. 
This method brought performance improvements in some cases, whereas it degraded performance in others. Thus, the experiments are excluded from the experimental results section.

\subsection{\textbf{Analyzing Hate Speech Category vs Strength}}
\label{sec:cat-str}

While a strict order of severity among different hate speech categories is not  accepted in the literature (e.g. the level of hate speech in a tweet in the “Symbolization'' category can be greater compared to one in the “Swearing'' category), we analyze the hate speech strength distribution in each category to better understand the problem and exploit the fact that the adopted hate speech categories show a progression of severity in order to motivate our 
ordinal assumption used in Section \ref{sec:no-clr}.

\begin{figure}
\label{fig:dists}
\centering
\includegraphics[width=1\linewidth]{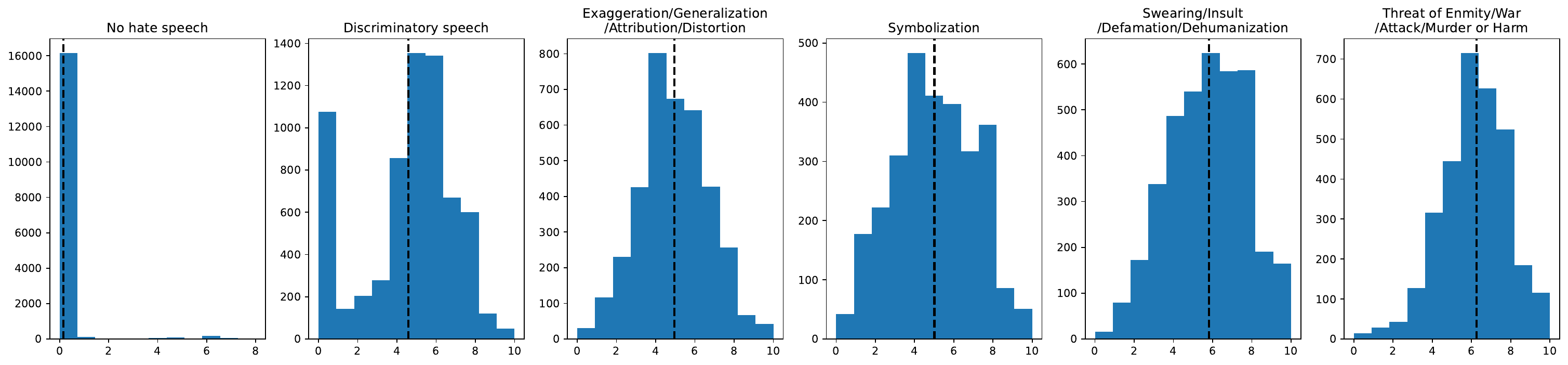}
\caption{Hate speech strength histograms for the Silver test set. Vertical dashed lines represent the mean of the distributions as [0.15, 4.59, 4.94, 5.01, 5.81, 6.26]}
   \label{fig:class_dist6} 
\end{figure}

As previously noted, during data collection and labeling, annotators were asked to select up to two speech categories per tweet, as well as a single hate speech strength value ranging from 0 to 10,
representing the severity of hate speech in the tweet. 
This category-strength relationship allows us to obtain the strength distributions for each class (if a tweet is labeled with two labels, it will contribute to to separate categories). 
Figure \ref{fig:class_dist6} illustrates these distributions within the original 6-class setting.


The first observation we can derive from the class-strength distributions is the considerable variability in the assigned strength scores. 
Specifically, for all classes except class-0 (“No Hate Speech''), the strength values span the entire range from 0 to 10. 
\textcolor{black}{This wide range of scores highlights the inherent subjectivity involved in assessing hate speech: while the guidelines tell annotators how to categorize each tweet, they themselves decide on the hate speech strength freely. }

We also note that the class “Exclusion/Discriminatory Discourse'' (class-1) 
has a significant number of samples labeled with a strength value of 0.
In other words, the annotator who follows the guideline feels that the tweet is discriminatory, but personally, they think it has a strength of 0. Indeed, whether or not classifying discriminatory speech as hate speech was one of the most debated points among annotators. 

More importantly, the mean strength scores for the categories show a strict increase from category-0 to category-5, with mean strengths of 0.15, 4.59, 4.94, 5.01, 5.81, and 6.26. This observation motivates our assumption of ordering the categories in increasing strength for suggesting label aggregation methods. 
When the categories are mapped to the $[0,5]$ range, the correlation with the strength values is highest according to Spearman’s rank correlation (0.8174), followed by Pearson’s correlation (0.7147), and Kendall’s tau (0.6784). The high correlation values show that the category labels can be seen as ordinal to a large extent.

\subsection{\textbf{Category Consolidation}}\vspace{0pt} 
\label{sec:mv-setting}

Multi-class hate speech classification typically uses between 4 and 6 categories. In order to evaluate our methodology with different numbers of categories, we describe how we obtain 4-class and 2-class categories from the original annotations.

\begin{figure}
    \centering
   \includegraphics[width=0.66\linewidth]{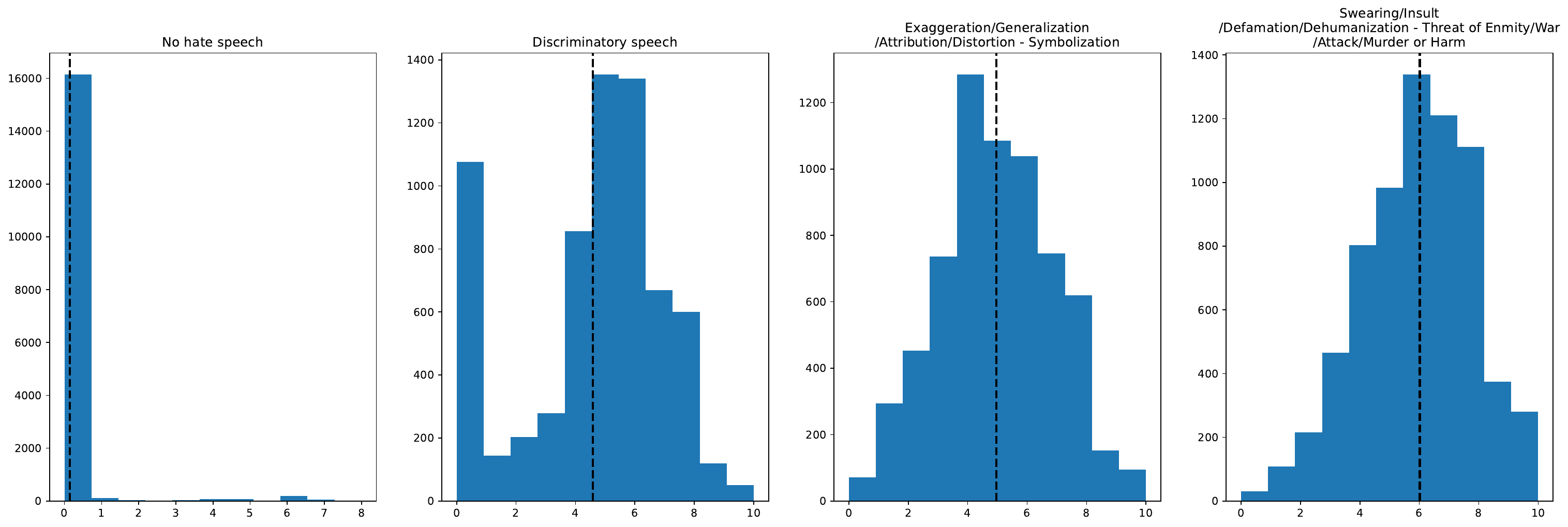}
   \caption{Strength distributions for the 4-class setting. Vertical dashed lines represent the mean of the distributions. Means are as follows: [0.15, 4.59, 4.97, 6.01]}
   \label{fig:class_dist4}
\end{figure}

In the analysis given in Section \ref{sec:cat-str}, the mean strengths for categories 2 and 3 were observed to be  similar (4.94 and 5.01), followed by the means of categories 4 and 5 (5.81 and 6.26). 
This observation supports the rationale for grouping classes 2 and 3 and classes 4 and 5 in the 4-class setting. 
Distributions for the resulting 4-class setting are improved in the sense that class means are more separated, as depicted in Figure \ref{fig:class_dist4}. 
Accordingly, we first map the six-class labels to the corresponding four-class labels, then calculate the counts of four-class labels before taking the majority vote:

\begin{equation}
c_l^{(4)}(x) = \sum_k \sum_{l'\in A_k(x)} \mathds{1} \left(l=r_4(l') \right)
\end{equation}

where $\mathds{1} \left(l=r_4(l') \right)$ is $1$ if $l=r_4(l')$, $0$ otherwise, and $r_4$ is the reduction mapping from the six-class setting to the four-class setting:
\begin{equation*}
r_4(l) = \begin{cases}
0 & l=0 \\
1 & l=1 \\
2 & l=2 \;\;\text{or}\;\; l=3 \\
3 & l=4 \;\;\text{or}\;\; l=5
\end{cases}
\end{equation*}

For the two-class setting, we apply the same label mapping strategy, where all hateful classes (1–5) are merged into a single positive class, while the non-hate class (0) remains unchanged and is kept as a separate category. Thus the mapping function is:
\begin{equation*}
r_2(l) = \begin{cases}
0 & l=0 \\
1 & \text{otherwise}
\end{cases}
\end{equation*}

Then the counts are calculated as follows:

\begin{equation}
c_l^{(2)}(x) = \sum_k \sum_{l'\in A_k(x)} \mathds{1} \left(l=r_2(l') \right)
\end{equation}

\subsection{Model Architecture and Training} \label{sec:model}

We design our classification model using transfer learning, incorporating a single layer on top of the BERT \citep{Devlin2019} 2019) encoder to predict hate speech categories or strengths. BERT, being a state-of-the-art model, has demonstrated success across numerous NLP tasks, including in our previous works \citep{Beyhan2022, Arin2023, Dehghan2024a, Dehghan2024b}, where it consistently achieved strong performance. Beyond classification tasks, BERT has also proven effective in semantic textual similarity \citep{Dehghan2022, Dehghan2023, Dehghan2025}, named entity recognition \citep{Liuner2021, Ullah2024}, question answering \citep{Duan2022, Mountantonakis2025, Weng2025}, and Sentiment analysis \citep{Ring2024, Awlla2025} .

We use BERTurk\footnote{https://huggingface.co/dbmdz/bert-base-turkish-uncased} checkpoint from Huggingface Transformer package and  apply fine-tuning for 15 epochs. We reserve 5\% of the training data as validation and select the model with the best validation performance --accuracy for the classification tasks. During training, we use a learning rate of $5 \times 10^{-6}$, batch-size of $16$ and apply weight-decay regularization with a parameter of $0.01$.
For the classification problem, we use the cross-entropy loss:
\begin{equation}
    L_{CE} = -\sum_{i=1}^{N} y_ilog(\hat y_i)
\end{equation}
where $y_i$ is the target value for the $i^{th}$ input and $\hat y_i$ is the prediction.

When we treated the problem as a regression problem, predicting the hate speech strength, we  trained the model using the mean of the annotated strength scores (ranging from 0 to 10) as the aggregate label, employing the Mean Squared Error (MSE) as the loss function. 
\begin{equation}
        L_{MSE} = \frac{1}{N}\sum_{i=1}^{N} (y_i- \hat y_i)^2
\end{equation}
where $y_i$ and $\hat y_i$ are the desired and predicted values, respectively.

\section{Experimental Results} \label{sec:experiment}

\subsection{Classification Experiments}
We conducted four classification experiments (E1-E4) 
focusing on how to handle annotator disagreements and evaluated the performance of the model on the designated test sets. 

For evaluating the models, we use two different test sets: \textit{Gold} and \textit{Silver}. The Gold test set contains only those samples where all annotators are in agreement, ensuring high reliability. There are  646/652/710 samples  in the Gold test set for 6-, 4-, and 2-class problems, respectively.
The Silver test set includes both the agreed-upon samples and those with disagreement among annotators but a clear majority vote (MV). This allows us to evaluate model performance on consensus data (Gold) versus a broader, more realistic set (Silver). There are  1,894/2,022/1,831 samples  in the Silver test set for 6-, 4-, and 2-class problems, respectively. 
Once the test sets were fixed for all classification experiments, we used the following experiments with different \textit{training sets:}

\vspace{3pt}
\textbf{Experiment 1:} We only kept the training samples for which there is consensus (i.e. all annotators have picked the same single category). 
Note that this approach filters out a large portion of the data, also considering that for full agreement, there must be a single selection from each annotator; thus, tweets involving multiple classes are also filtered out. 
Furthermore, as annotators generally agree about the no hate speech class, the resulting set is very unbalanced.
As a result, there are 2,580/2,606/2,786 training samples for 6-, 4- and 2-class tasks. 

\textbf{Experiment 2:} In addition to the consensus samples, we also kept the training samples where the majority vote is clear (see \ref{sec:aggreg}).
This increases both the total number of tweets kept and  the classes are more balanced. 
\textcolor{black}{As a result, there are more than 7,300 samples in this category.
}

\textbf{Experiment 3:} In this experiment, we used all 8,820 samples for training, but the label distribution varied by aggregation strategy and number of classes.
The proportion of the samples with disagreement is approximately 
16.4\%, 15\%, and 8\%  for the 6-class, 4-class, and 2-class problems, respectively.

\begin{table}[tb]
\centering
\small
\setlength{\tabcolsep}{7pt}
\caption{Results of hate speech classification model (BERTurk) for different strategies to handle disagreement among annotators on 2-class, 4-class, and 6-class problems on Gold (Consensus samples) and Silver (Consensus and Clear-majority samples) test sets. Bold-underlined and Bold values indicate best results for the Gold and Silver test sets respectively, in each experiment.}
\label{tab:classification-result}
\begin{tabular}{|l|l|l|cccccc|} 
\cline{1-9}

\multicolumn{1}{|c|}{\multirow{3}{*}{Exp.}} & \multicolumn{1}{c|}{\multirow{3}{*}{Training Set - Majority Voting Strategy}}                                                  & \multicolumn{1}{c|}{\multirow{3}{*}{\begin{tabular}[c]{@{}c@{}}Test\\ Set\end{tabular}}} & \multicolumn{6}{c|}{Classification Problem}                                                                                                        \\ \cline{4-9} 
\multicolumn{1}{|c|}{}                      & \multicolumn{1}{c|}{}                                                                                                          & \multicolumn{1}{c|}{}                                                                    & \multicolumn{2}{c|}{6-class}                     & \multicolumn{2}{c|}{4-class}                     & \multicolumn{2}{c|}{2-class}                 \\ \cline{4-9} 
\multicolumn{1}{|c|}{}                      & \multicolumn{1}{c|}{}                                                                                                          & \multicolumn{1}{c|}{}                                                                    & M-F1                 & \multicolumn{1}{c|}{Acc.} & M-F1                 & \multicolumn{1}{c|}{Acc.} & M-F1                 & Acc.                  \\ \cline{1-9} 

\multirow{2}{*}{E1}                         & \multirow{2}{*}{Consensus samples}                                                                                            & Gold                                                                                     & \underline{\textbf{71.59}}                    & \multicolumn{1}{c|}{\underline{\textbf{97.67}}}    & \underline{\textbf{82.98}}                   & \multicolumn{1}{c|}{\underline{\textbf{97.54}}}    & 88.02                    & 92.81                     \\
                                            &                                                                                                                                & Silver                                                                                   & 48.66                    & \multicolumn{1}{c|}{76.62}    & 56.50                    & \multicolumn{1}{c|}{73.81}    & 73.20                    & 74.67                     \\ \cline{1-9} 

\multirow{2}{*}{E2}                         & \multirow{2}{*}{Consensus + Clear majority samples}                                                                           & Gold                                                                                     & 65.51                    & \multicolumn{1}{c|}{94.73}    & 76.60                    & \multicolumn{1}{c|}{94.01}    & 85.13                    & 89.57                     \\
                                            &                                                                                                                                & Silver                                                                                   & 62.23                    & \multicolumn{1}{c|}{79.51}    & 71.01                    & \multicolumn{1}{c|}{79.14}    & 83.18                    & 83.23                     \\ \cline{1-9}

\multirow{8}{*}{E3}                         & \multirow{2}{*}{All samples - Max Aggregation}                                                                                             & Gold                                                                                     & 60.93 & \multicolumn{1}{l|}{91.79} & 75.71 & \multicolumn{1}{l|}{93.55} & 83.47 & \multicolumn{1}{l|}{87.74} \\
                                            &                                                                                                                                & Silver                                                                                   & 61.23                    & \multicolumn{1}{c|}{77.88}    & \textbf{72.92}                    & \multicolumn{1}{c|}{\textbf{79.61}}    & \textbf{83.63}                   & \textbf{83.63}                     \\ \cline{2-9} 

                                            & \multirow{2}{*}{All samples - Mean Aggregation}                                                                                            & Gold                                                                                     & 63.91 & \multicolumn{1}{l|}{89.93} & 76.74 & \multicolumn{1}{l|}{93.55} & 90.56 & \multicolumn{1}{l|}{93.94} \\
                                            &                                                                                                                                & Silver                                                                                   & 60.47                    & \multicolumn{1}{c|}{75.86}    & 71.55                    & \multicolumn{1}{c|}{78.56}    & 82.23                    & 82.54                     \\ \cline{2-9} 

& \multirow{2}{*}{All samples - Min Aggregation}                                                                                              & Gold                                                                                     & 64.33 & \multicolumn{1}{l|}{95.35} & 80.74 & \multicolumn{1}{l|}{95.55} & \underline{\textbf{90.56}} & \multicolumn{1}{l|}{\underline{\textbf{93.94}}} \\          
                                            &                                                                                                                                & Silver                                                                                   & \textbf{62.99}                    & \multicolumn{1}{c|}{\textbf{80.22}}    & 69.25                    & \multicolumn{1}{c|}{78.29}    & 82.23                    & 82.54                     \\ \cline{2-9} 
                                            & \multirow{2}{*}{All samples - Random Aggregation}                                                                                          & Gold                                                                                     & 63.97 & \multicolumn{1}{l|}{94.89} & 73.20 & \multicolumn{1}{l|}{95.39} & 84.11 & \multicolumn{1}{l|}{88.59} \\
                                            &                                                                                                                                & Silver                                                                                   & 59.80                    & \multicolumn{1}{c|}{79.24}    & 64.36                    & \multicolumn{1}{c|}{76.24}    & 81.40                    & 81.45                     \\ \cline{1-9}

\multirow{2}{*}{E4}                         & \multirow{2}{*}{\begin{tabular}[c]{@{}l@{}}Ensemble Model (Min for Lenient Model \\ and Max for Sensitive Model)\end{tabular}} & Gold                                                                                     & 64.46 & \multicolumn{1}{l|}{91.33} & 72.85 & \multicolumn{1}{l|}{92.63} & 90.56 & \multicolumn{1}{l|}{93.94} \\
                                            &                                                                                                                                & Silver                                                                                   & 59.26                    & \multicolumn{1}{c|}{76.18}    & 64.27                    & \multicolumn{1}{c|}{75.29}    & 82.08                    & 82.39                     \\ \cline{1-9}

\end{tabular}
\end{table}

\textbf{Experiment 4:} \textcolor{black}{In this experiment, we apply an averaging ensemble, given two classifier models: Lenient and Sensitive. These classifiers are trained using  all the data as in Experiment 3. For the Lenient classifier, we adopt the minimum label value among annotators, representing the most lenient (not sensitive) interpretation. For the Sensitive classifier, we adopt the maximum label value, representing the most sensitive interpretation. The final prediction is obtained by averaging the outputs of both classifiers, aiming to balance between lenient and sensitive judgments in ambiguous cases.}

\subsection{Observations on Classification Results} \label{sec:result}
The results corresponding to Experiments 1-4 are given in Table \ref{tab:classification-result}.
The experimental results demonstrate that the manner in which annotator disagreement is handled can significantly impact model performance in hate speech classification. 

\textbf{Results on the Gold Test Set:} 
\textcolor{black}{
The model trained with only consensus samples achieved the best on the 6- and 4-class classification tasks on the Gold test (and second best on the 2-class task),} indicating that the training set consisting of fully agreed upon samples is sufficient to learn the relatively simple problem of classifying fully agreed upon samples.

\textbf{Gold vs Silver Test Sets:}
When comparing the results obtained for Gold versus Silver test sets  in Experiments E1 or E2 , we observe that the Silver test results are significantly and consistently lower (over 20\% points in F1 score and accuracy) in each experiment.
The noticeable drop in performance between the Gold and Silver test sets  across all experiments underscores the complexities introduced when deciding on tweets for which annotators do not have a consensus.
While this result is not surprising, we believe this is the first comparison of how much discarding samples without full agreement affects performance.

\textbf{Aggregation strategy when there is no clear majority:} 
In Experiment 3, all training samples are used, but with a different majority voting strategy to derive their aggregate label (Min, Max, etc.).
%
%
\textcolor{black}{
The results do not identify a single aggregation strategy that is consistently optimal across all classification granularities. 
For the coarser 2- and 4-class settings, Max strategy generally achieves the strongest performance, suggesting that assigning the more severe label can improve sensitivity to hate speech. In contrast, for the finer 6-class setting, Min strategy performs best on the Silver test set, possibly because it produces labels that are more consistent with the underlying class distribution (see Table \ref{tab:error-analysis-dikey-hepsi}). 
However, when  performance is averaged across the three classification tasks, the Max aggregation strategy performs best on the Silver test set (average F1 and accuracy of 72.59\% and 80.37\%), with Experiment 2 yielding comparable results (average F1 and accuracy of 72.14\% and 80.63\%). 
In real-world settings where ambiguous cases are also present at test time, aggregation-based training strategies may offer additional benefits, a possibility that warrants further investigation.}
%

\textbf{Value of Noisy Training Samples}: For the 2-class problem, the Min strategy used with all the training samples achieved the highest M-F1 (90.56\%) and accuracy scores (93.94\%) on the Gold set, surpassing the results obtained in Experiment E1. 
Furthermore, on the larger and more realistic Silver test set, E1 results were  consistently outperformed by those of experiments E2-E4. These observations  suggest that using only the “cleanest'' training data is not always sufficient for the more realistic setting. 
On the contrary, incorporating a broader and more diverse set of training examples—including samples with clear and unclear majority voting—can enhance the model's ability to generalize, especially when disagreement is handled effectively through strategies such as Min or Max depending on the task resolution. Therefore, a carefully curated combination of clean and ambiguous data can be more beneficial than relying exclusively on fully agreed annotations.

\textbf{Sensitive versus Lenient Models}:
In Experiment E4, we evaluated an ensemble classifier that combined two classifiers: the lenient model trained with the Min strategy and the sensitive model trained with the Max strategy. Our motivation in this experiment was to simulate the range of annotators who could take a sensitive or lenient stance.
This hybrid model maintained high performance in the 2-class setting (M-F1: 90.56\%), but did not surpass the simpler E3 strategies in 4-class or 6-class tasks.


\begin{table}
\setlength{\tabcolsep}{4.7pt}
\centering
\scriptsize
\caption{Confusion matrices and class-based accuracies for the 6-, 4-, 2-class problems with the more realistic Silver test set in Experiment 3. The first row shows results for Max aggregation and the second row show results for Min aggregation.}
\label{tab:error-analysis-dikey-hepsi}
\begin{tabular}{llcccccccllccllllllll}
\cline{1-21}
                                                
                                                &                        & \multicolumn{6}{c}{Predicted Label}                                                                                                                                                                                & \multicolumn{1}{l}{}                                        &                             &                        & \multicolumn{4}{c}{Predicted Label}                                                                                                            &                                                                                &                             &                        & \multicolumn{2}{c}{Predicted Label}                                   &                                                                                \\
                                                &                        & 0                                  & 1                                 & 2                                & 3                                & 4                                & 5                                & \begin{tabular}[c]{@{}c@{}}Acc.\\ (per class)\end{tabular}  &                             &                        & 0                                  & 1                                 & \multicolumn{1}{c}{2}             & \multicolumn{1}{c}{3}             & \multicolumn{1}{c}{\begin{tabular}[c]{@{}c@{}}Acc.\\ (per class)\end{tabular}} &                             &                        & \multicolumn{1}{c}{0}             & \multicolumn{1}{c}{1}             & \multicolumn{1}{c}{\begin{tabular}[c]{@{}c@{}}Acc.\\ (per class)\end{tabular}} \\ \cline{3-8} \cline{12-15} \cline{19-20}
\multicolumn{1}{c}{\multirow{6}{*}{\rotatebox{90}{True Label}}} & \multicolumn{1}{c|}{0} & \multicolumn{1}{c|}{\textbf{980}}  & \multicolumn{1}{c|}{22}           & \multicolumn{1}{c|}{24}          & \multicolumn{1}{c|}{22}          & \multicolumn{1}{c|}{42}          & \multicolumn{1}{c|}{39}          & 86.8                                                        & \multirow{4}{*}{\rotatebox{90}{True Label}} & \multicolumn{1}{l|}{0} & \multicolumn{1}{c|}{\textbf{999}}  & \multicolumn{1}{c|}{20}           & \multicolumn{1}{c|}{49}           & \multicolumn{1}{c|}{54}           & \multicolumn{1}{c}{89.0}                                                       & \multirow{4}{*}{\rotatebox{90}{True Label}} & \multicolumn{1}{l|}{0} & \multicolumn{1}{c|}{\textbf{845}} & \multicolumn{1}{c|}{205}          & \multicolumn{1}{c}{80.5}                                                       \\ \cline{3-8} \cline{12-15} \cline{19-20}
\multicolumn{1}{c}{}                            & \multicolumn{1}{c|}{1} & \multicolumn{1}{c|}{9}             & \multicolumn{1}{c|}{\textbf{199}} & \multicolumn{1}{c|}{5}           & \multicolumn{1}{c|}{5}           & \multicolumn{1}{c|}{12}          & \multicolumn{1}{c|}{2}           & 85.7                                                        &                             & \multicolumn{1}{l|}{1} & \multicolumn{1}{c|}{7}             & \multicolumn{1}{c|}{\textbf{180}} & \multicolumn{1}{c|}{8}            & \multicolumn{1}{c|}{11}           & \multicolumn{1}{c}{87.4}                                                       &                             & \multicolumn{1}{l|}{1} & \multicolumn{1}{c|}{126}          & \multicolumn{1}{c|}{\textbf{846}} & \multicolumn{1}{c}{87.0}                                                       \\ \cline{3-8} \cline{12-15} \cline{19-20}
\multicolumn{1}{c}{}                            & \multicolumn{1}{c|}{2} & \multicolumn{1}{c|}{33}            & \multicolumn{1}{c|}{8}            & \multicolumn{1}{c|}{\textbf{28}} & \multicolumn{1}{c|}{6}           & \multicolumn{1}{c|}{21}          & \multicolumn{1}{c|}{5}           & 27.7                                                        &                             & \multicolumn{1}{l|}{2} & \multicolumn{1}{c|}{64}            & \multicolumn{1}{c|}{8}            & \multicolumn{1}{c|}{\textbf{123}} & \multicolumn{1}{c|}{48}           & \multicolumn{1}{c}{50.6}                                                       &                             &                        & M-F1 :                            & \multicolumn{1}{c}{83.63}         &                                                                                \\ \cline{3-8} \cline{12-15}
\multicolumn{1}{c}{}                            & \multicolumn{1}{c|}{3} & \multicolumn{1}{c|}{16}            & \multicolumn{1}{c|}{1}            & \multicolumn{1}{c|}{3}           & \multicolumn{1}{c|}{\textbf{40}} & \multicolumn{1}{c|}{20}          & \multicolumn{1}{c|}{3}           & 48.2                                                        &                             & \multicolumn{1}{l|}{3} & \multicolumn{1}{c|}{75}            & \multicolumn{1}{c|}{5}            & \multicolumn{1}{c|}{37}           & \multicolumn{1}{c|}{\textbf{206}} & \multicolumn{1}{c}{63.8}                                                       &                             &                        & Acc. :                            & \multicolumn{1}{c}{83.63}         &                                                                                \\ \cline{3-8} \cline{12-15}
\multicolumn{1}{c}{}                            & \multicolumn{1}{c|}{4} & \multicolumn{1}{c|}{24}            & \multicolumn{1}{c|}{2}            & \multicolumn{1}{c|}{6}           & \multicolumn{1}{c|}{12}          & \multicolumn{1}{c|}{\textbf{92}} & \multicolumn{1}{c|}{12}          & 62.2                                                        &                             &                        & \multicolumn{1}{l}{M-F1 :}         & 72.92                             &                                   &                                   &                                                                                &                             &                        &                                   &                                   &                                                                                \\ \cline{3-8}
\multicolumn{1}{c}{}                            & \multicolumn{1}{c|}{5} & \multicolumn{1}{c|}{32}            & \multicolumn{1}{c|}{3}            & \multicolumn{1}{c|}{0}           & \multicolumn{1}{c|}{3}           & \multicolumn{1}{c|}{13}          & \multicolumn{1}{c|}{\textbf{87}} & 63.0                                                        &                             &                        & \multicolumn{1}{l}{Acc. :}         & 79.61                             &                                   &                                   &                                                                                &                             &                        &                                   &                                   &                                                                                \\ \cline{3-8}
                                                &                        & \multicolumn{1}{l}{M-F1 :}         & 61.23                             & \multicolumn{1}{l}{}             & \multicolumn{1}{l}{}             & \multicolumn{1}{l}{}             & \multicolumn{1}{l}{}             & \multicolumn{1}{l}{}                                        &                             &                        & \multicolumn{1}{l}{}               & \multicolumn{1}{l}{}              &                                   &                                   &                                                                                &                             &                        &                                   &                                   &                                                                                \\
                                                &                        & \multicolumn{1}{l}{Acc. :}         & 77.88                             & \multicolumn{1}{l}{}             & \multicolumn{1}{l}{}             & \multicolumn{1}{l}{}             & \multicolumn{1}{l}{}             & \multicolumn{1}{l}{}                                        &                             &                        & \multicolumn{1}{l}{}               & \multicolumn{1}{l}{}              &                                   &                                   &                                                                                &                             &                        &                                   &                                   &                                                                                \\
                                                &                        & \multicolumn{1}{l}{}               & \multicolumn{1}{l}{}              & \multicolumn{1}{l}{}             & \multicolumn{1}{l}{}             & \multicolumn{1}{l}{}             & \multicolumn{1}{l}{}             & \multicolumn{1}{l}{}                                        &                             &                        & \multicolumn{1}{l}{}               & \multicolumn{1}{l}{}              &                                   &                                   &                                                                                &                             &                        &                                   &                                   &                                                                                \\
                                                &                        & \multicolumn{1}{l}{}               & \multicolumn{1}{l}{}              & \multicolumn{1}{l}{}             & \multicolumn{1}{l}{}             & \multicolumn{1}{l}{}             & \multicolumn{1}{l}{}             & \multicolumn{1}{l}{}                                        &                             &                        & \multicolumn{1}{l}{}               & \multicolumn{1}{l}{}              &                                   &                                   &                                                                                &                             &                        &                                   &                                   &                                                                                \\
                                                &                        & \multicolumn{6}{c}{Predicted Label}                                                                                                                                                                                &                                                             &                             &                        & \multicolumn{4}{c}{Predicted Label}                                                                                                            & \multicolumn{1}{c}{}                                                           & \multicolumn{1}{c}{}        &                        & \multicolumn{2}{c}{Predicted Label}                                   & \multicolumn{1}{c}{}                                                           \\
                                                &                        & 0                                  & 1                                 & 2                                & 3                                & 4                                & 5                                & \begin{tabular}[c]{@{}c@{}}Acc. \\ (per class)\end{tabular} &                             &                        & 0                                  & 1                                 & \multicolumn{1}{c}{2}             & \multicolumn{1}{c}{3}             & \multicolumn{1}{c}{\begin{tabular}[c]{@{}c@{}}Acc.\\ (per class)\end{tabular}} &                             &                        & \multicolumn{1}{c}{0}             & \multicolumn{1}{c}{1}             & \multicolumn{1}{c}{\begin{tabular}[c]{@{}c@{}}Acc.\\ (per class)\end{tabular}} \\ \cline{3-8} \cline{12-15} \cline{19-20}
\multicolumn{1}{c}{\multirow{6}{*}{\rotatebox{90}{True Label}}} & \multicolumn{1}{c|}{0} & \multicolumn{1}{c|}{\textbf{1058}} & \multicolumn{1}{c|}{17}           & \multicolumn{1}{c|}{10}          & \multicolumn{1}{c|}{4}           & \multicolumn{1}{c|}{15}          & \multicolumn{1}{c|}{25}          & 93.7                                                        & \multirow{4}{*}{\rotatebox{90}{True Label}} & \multicolumn{1}{l|}{0} & \multicolumn{1}{c|}{\textbf{1051}} & \multicolumn{1}{c|}{17}           & \multicolumn{1}{c|}{26}           & \multicolumn{1}{c|}{28}           & \multicolumn{1}{c}{93.7}                                                       & \multirow{2}{*}{\rotatebox{90}{True Label}} & \multicolumn{1}{l|}{0} & \multicolumn{1}{c|}{\textbf{967}} & \multicolumn{1}{c|}{83}           & \multicolumn{1}{c}{92.1}                                                       \\ \cline{3-8} \cline{12-15} \cline{19-20}
\multicolumn{1}{c}{}                            & \multicolumn{1}{c|}{1} & \multicolumn{1}{c|}{20}            & \multicolumn{1}{c|}{\textbf{202}} & \multicolumn{1}{c|}{3}           & \multicolumn{1}{c|}{2}           & \multicolumn{1}{c|}{4}           & \multicolumn{1}{c|}{1}           & 87.1                                                        &                             & \multicolumn{1}{l|}{1} & \multicolumn{1}{c|}{13}            & \multicolumn{1}{c|}{\textbf{187}} & \multicolumn{1}{c|}{4}            & \multicolumn{1}{c|}{2}            & \multicolumn{1}{c}{90.8}                                                       &                             & \multicolumn{1}{l|}{1} & \multicolumn{1}{c|}{270}          & \multicolumn{1}{c|}{\textbf{702}} & \multicolumn{1}{c}{72.2}                                                       \\ \cline{3-8} \cline{12-15} \cline{19-20}
\multicolumn{1}{c}{}                            & \multicolumn{1}{c|}{2} & \multicolumn{1}{c|}{54}            & \multicolumn{1}{c|}{6}            & \multicolumn{1}{c|}{\textbf{30}} & \multicolumn{1}{c|}{4}           & \multicolumn{1}{c|}{5}           & \multicolumn{1}{c|}{2}           & 29.7                                                        &                             & \multicolumn{1}{l|}{2} & \multicolumn{1}{c|}{114}           & \multicolumn{1}{c|}{8}            & \multicolumn{1}{c|}{\textbf{97}}  & \multicolumn{1}{c|}{24}           & \multicolumn{1}{c}{39.9}                                                       & \multicolumn{1}{c}{}        &                        & M-F1 :                            & \multicolumn{1}{c}{82.23}         &                                                                                \\ \cline{3-8} \cline{12-15}
\multicolumn{1}{c}{}                            & \multicolumn{1}{c|}{3} & \multicolumn{1}{c|}{25}            & \multicolumn{1}{c|}{0}            & \multicolumn{1}{c|}{5}           & \multicolumn{1}{c|}{\textbf{46}} & \multicolumn{1}{c|}{5}           & \multicolumn{1}{c|}{2}           & 55.4                                                        &                             & \multicolumn{1}{l|}{3} & \multicolumn{1}{c|}{123}           & \multicolumn{1}{c|}{17}           & \multicolumn{1}{c|}{35}           & \multicolumn{1}{c|}{\textbf{148}} & \multicolumn{1}{c}{45.8}                                                       & \multicolumn{1}{c}{}        &                        & Acc. :                            & \multicolumn{1}{c}{82.54}         &                                                                                \\ \cline{3-8} \cline{12-15}
\multicolumn{1}{c}{}                            & \multicolumn{1}{c|}{4} & \multicolumn{1}{c|}{46}            & \multicolumn{1}{c|}{5}            & \multicolumn{1}{c|}{2}           & \multicolumn{1}{c|}{18}          & \multicolumn{1}{c|}{\textbf{60}} & \multicolumn{1}{c|}{17}          & 40.5                                                        & \multicolumn{1}{c}{}        &                        & \multicolumn{1}{l}{M-F1 :}         & 69.25                             &                                   &                                   &                                                                                &                             &                        &                                   &                                   &                                                                                \\ \cline{3-8}
\multicolumn{1}{c}{}                            & \multicolumn{1}{c|}{5} & \multicolumn{1}{c|}{48}            & \multicolumn{1}{c|}{3}            & \multicolumn{1}{c|}{3}           & \multicolumn{1}{c|}{2}           & \multicolumn{1}{c|}{9}           & \multicolumn{1}{c|}{\textbf{73}} & 52.9                                                        & \multicolumn{1}{c}{}        &                        & \multicolumn{1}{l}{Acc. :}         & 78.29                             &                                   &                                   &                                                                                &                             &                        &                                   &                                   &                                                                                \\ \cline{3-8}
                                                &                        & \multicolumn{1}{l}{M-F1 :}         & 62.99                             & \multicolumn{1}{l}{}             & \multicolumn{1}{l}{}             & \multicolumn{1}{l}{}             & \multicolumn{1}{l}{}             & \multicolumn{1}{l}{}                                        &                             &                        & \multicolumn{1}{l}{}               & \multicolumn{1}{l}{}              &                                   &                                   &                                                                                &                             &                        &                                   &                                   &                                                                                \\
                                                &                        & \multicolumn{1}{l}{Acc. :}         & 80.22                             & \multicolumn{1}{l}{}             & \multicolumn{1}{l}{}             & \multicolumn{1}{l}{}             & \multicolumn{1}{l}{}             & \multicolumn{1}{l}{}                                        &                             &                        & \multicolumn{1}{l}{}               & \multicolumn{1}{l}{}              &                                   &                                   &                                                                                &                             &                        &                                   &                                   &                                                                                \\   \cline{1-21}

\end{tabular}
\end{table}

\subsection{Regression Model Using Hate Speech Strength } \label{sec:regresult}
In Experiment E5, we trained a model to predict the average hate speech strength.
For each tweet, we calculated the mean of the annotated strength scores as the label (ranging from 0 to 10) and trained a regression model using the  Mean Squared Error (MSE) loss function.

\textcolor{black}{In addition to measuring the regression performance using Root Mean Squared Error (RMSE) metric, we obtained a binary classifier by applying a threshold to the predicted scores.  
Instead of using a fixed threshold of 0.5, we selected an optimal threshold that maximized accuracy on the validation set. 
To assess the {binary classification performance} based on the perceived strength, we derived the ground-truth labels from the mean perceived  scores: tweets with a mean score above 0.5 were labeled as “Hate'', while those with a score of 0.5 or below were labeled as “No-hate''. 
}

This experiment was conducted with different subsets of the data to compare performance results from subsets with full annotator agreement to those including all data, to examine the impact of excluding data points lacking consensus among annotators. 
To identify cases of agreement among annotators, we first converted the strength annotations into auxiliary binary values: a score of 0 was mapped to 0, while scores in the range of [1, 10] were mapped to 1. We then selected tweets for which all annotators assigned the same binary value to find the consensus subset. 
``Agreements'' data is the subset of data points for which all annotators agree on the thresholded binary class, i.e., all zeros or all strictly positive (in the range [1-10]); whereas ``All'' data does not exclude any data points.

We evaluated the trained regression models in a binary classification setting, reporting performance using F1 and accuracy metrics, in addition to the Root Mean Squared Error (RMSE).
The results of the regression experiments are presented in Table \ref{tab:regression}. 

When trained and tested with the samples where there is agreement among all annotators, the model shows very high performance with a RMSE of 1.64 and an accuracy of 89.80\% (top row). However, when tested with  all test data (second row), the accuracy drops by more than 10\% points.

As with the classification experiments, the best results on the unfiltered (all) test set  are obtained when using all training samples (third row). 
The accuracy of 79.42\% can be compared to the best result obtained in Experiment E3 with Max aggregation (89.43\% accuracy) and indicates that the perceived hate speech strength works well as a proxy for detailed annotation guidelines. 

Similar to classification results, the model  performs better when trained with all data (79.42\% vs 78.91\% accuracy). This result suggests not filtering out any data in search of annotator agreement samples.

\begin{table}
\centering
\small
\setlength{\tabcolsep}{11pt}
\caption{Regression and binary classification using the hate speech strength on a 0-10 scale. The best results are shown in bold. }
\begin{tabular}{|c|l|c|c|c|c|c|}
\cline{1-7}
Exp.                & Modeling Type              & Train-data & Test-data  & RMSE & 2-class M-F1  & 2-class Acc.  \\ \cline{1-7}
\multirow{3}{*}{E5} & \multirow{3}{*}{Regression} & Agreements & Agreements & 1.64 & 89.57 & 89.80 \\
                    &                             & Agreements & All        & 1.92 & 77.88 & 78.91 \\
                    &                             & All        & All        & 1.60 & \textbf{78.44} & \textbf{79.42} \\ \cline{1-7}
\end{tabular}
\label{tab:regression}
\end{table}

\begin{table}
\centering
\small
\setlength{\tabcolsep}{8.8pt}
\caption{Literature results for Turkish hate speech classification. The best results are shown in bold.}
\label{tab:comparison}
\begin{tabular}{|l|cccccccc|}
\cline{1-9}
\multicolumn{1}{|c|}{\multirow{3}{*}{Model}} & \multicolumn{8}{c|}{Classification Problem}                                                                                                 \\ \cline{2-9} 
\multicolumn{1}{|c|}{}                       & \multicolumn{2}{c|}{2-class}       & \multicolumn{2}{c|}{4-class}       & \multicolumn{2}{c|}{5-class}       & \multicolumn{2}{c|}{6-class} \\ \cline{2-9} 

\multicolumn{1}{|c|}{}                       & M-F1  & \multicolumn{1}{c|}{Acc.}  & M-F1  & \multicolumn{1}{c|}{Acc.}  & M-F1  & \multicolumn{1}{c|}{Acc.}  & M-F1    & Acc.   

\\ \cline{1-9}

\citet{Beyhan2022}                          & 65.54 & \multicolumn{1}{c|}{71.06} & -     & \multicolumn{1}{c|}{-}     & 60.98 & \multicolumn{1}{c|}{71.74} & -             & -            \\ \cline{1-9}

\citet{Arin2023}      & 76.87 & \multicolumn{1}{c|}{-} & -     & \multicolumn{1}{c|}{-}     & 57.58 & \multicolumn{1}{c|}{-}     & -             & -            \\ \cline{1-9}

\citet{Uludogan2024a}       & 69.64 & \multicolumn{1}{c|}{-}     & -     & \multicolumn{1}{c|}{-}     & -     & \multicolumn{1}{c|}{-}     & -             & -            \\ \cline{1-9}

\citet{Dehghan2024b}                        & 79.49 & \multicolumn{1}{c|}{83.81} & -     & \multicolumn{1}{c|}{-}     & 54.99 & \multicolumn{1}{c|}{80.56} & -             & -            \\ \cline{1-9}

\makecell[l]{Our best results on the Silver \\ test set (Exp. 3 - Max Strategy)} 
& \textbf{83.63} 
& \multicolumn{1}{c|}{\textbf{83.63}} 
& 72.92 
& \multicolumn{1}{c|}{79.61} 
& -     
& \multicolumn{1}{c|}{-}     
& 61.23         
& 77.88       
\\ \cline{1-9}
\end{tabular}
\end{table}



\subsection{Comparison to Literature} \label{sec:comparison}

We have compiled various literature results on hate speech classification in Turkish, as shown in Table \ref{tab:comparison}. 
While our performance results are not directly comparable to any results in the literature, two competitions on the topic used earlier versions of this dataset. The winner of the  SIU2023-NST competition \citep{Arin2023} obtained a 76.87\% F1 score on the 2-class classification for a subset of the Refugee dataset. 
Similarly, in the HSD-2Lang competition, \citep{Uludogan2024a} achieved a macro F1 score of 69.64\%, while \citep{Dehghan2024b} attained a macro F1 score of 79.49\%, both evaluated on a combined test set of three topics (Anti-Refugees, Israel-Palestine conflict, and Anti Greek sentiment in Turkey).

\textcolor{black}{
For comparison, we use our best results on the Silver test set from  Experiment 3 (using Max aggregation).}
Our model outperformed all others, achieving a macro F1 of 83.63\% on an extended version of the same test set for 2-class problem.


\section{Conclusion and Future Work} \label{sec:conclusion}

Disagreements in hate speech annotation are often overlooked in the literature. Our work aims to highlight the problem and describe and evaluate alternatives to determine the aggregate label, in cases of multi-annotator and also multi-label annotation process.

\textcolor{black}{
Our experiments demonstrate that training exclusively on fully agreed-upon annotations (Experiment E1) offers strong baseline performance on the consensus samples (Gold test set); but it performs significantly worse on the larger and more realistic Silver set (consensus and clear-majority samples).
When performance is averaged across the three classification tasks, the Max aggregation strategy in Experiment E3 performs best on the Silver test set, followed closely by Experiment E2, which is trained using only consensus and clear-majority samples.}


\textcolor{black}{
Furthermore, although the regression-based results using the perceived hate speech strength (Experiment E5) are lower than those obtained with categorical labels (79.42\% vs 83.63\% for the 2-class classification task), we believe that perceived strength annotations offer a practical alternative for large-scale data collection. Annotators can assign a strength score much more quickly than selecting among multiple fine-grained categories, reducing both annotation effort and the need for detailed guidelines. Such labels could therefore be collected at scale and used as coarse supervisory signals, for example during pretraining or as an initial annotation stage prior to more detailed categorization.}


As future work, it would be valuable to explore the impact of annotators’ gender, demographic backgrounds, ethnicity, education level, personal outlook (optimistic or pessimistic), and personal biases on the quality of hate speech annotations. By examining how these factors influence the perception and labeling of hate speech, researchers could develop more comprehensive annotation guidelines and methods that account for these variations. This could lead to even more reliable datasets, ultimately enhancing the accuracy of hate speech detection models.

\vspace{12pt}
\paragraph{\normalfont{\textbf{Acknowledgments}}}
This article was produced within the scope of the project “Utilizing Digital Technology for Social Cohesion, Positive Messaging and Peace by Boosting Collaboration, Exchange and Solidarity'' (EuropeAid/170389/DD/ACT/Multi), implemented by the Hrant Dink Foundation in partnership with Sabancı University and Boğaziçi University, and supported by the European Union and the Friedrich Naumann Foundation. The implementing parties are solely responsible for the content of this publication, and the views expressed herein do not necessarily reflect those of the supporters.

\paragraph{\normalfont{\textbf{Author Contributions}}}
All authors contributed to the conception and design of the study.
Somaiyeh Dehghan and Mehmet Umut Sen jointly developed the methodology, implemented the classification and regression models, and conducted the related experiments. Somaiyeh Dehghan wrote the main parts of the manuscript, while Mehmet Umut Sen authored the sections on regression. Berrin Yanikoglu originated the research idea, led the project, and revised the manuscript. All authors read and approved the final manuscript.


\paragraph{\normalfont{\textbf{Funding Statement}}} 
This research was funded by the European Union and the Friedrich Naumann Foundation.


\paragraph{\normalfont{\textbf{Data Availability Statement}}}
The data supporting the findings of this study are not publicly available. They can be made available by the corresponding author for scientific or research purposes.

\paragraph{\normalfont{\textbf{Data Annotation Guideline Availability}}} The English and Turkish guidelines are publicly available\footref{guideurl_1}\textsuperscript{,}\footref{guideurl_2}.

\paragraph{\normalfont{\textbf{Competing Interests}}}
The authors have no relevant financial or non-financial interests to disclose. 


\paragraph{\normalfont{\textbf{Declaration of AI-Assisted Proofreading}}} Authors used AI to assist with proofreading the manuscript. All AI-generated suggestions were carefully reviewed and edited as necessary, and the authors take full responsibility for the final content of the published article.




\end{document}